\pdfoutput=1

\documentclass[conference]{IEEEtran}
\IEEEoverridecommandlockouts

\usepackage{cite}
\usepackage{amsmath,amssymb,amsfonts}
\usepackage{algorithmic}
\usepackage{graphicx}
\usepackage{textcomp}
\usepackage{xcolor}
\usepackage{wrapfig}
\usepackage{amsmath}
\usepackage{floatrow}
\usepackage{graphicx}
\usepackage{multirow}
\usepackage{booktabs}
\usepackage{subcaption}
\usepackage{cite}
\usepackage{verbatim}
\graphicspath{ {images/} }
\newtheorem{Pro}{Problem}

\usepackage{times}  
\usepackage[T1]{fontenc}  
\def\BibTeX{{\rm B\kern-.05em{\sc i\kern-.025em b}\kern-.08em
    T\kern-.1667em\lower.7ex\hbox{E}\kern-.125emX}}
\begin{document}

\title{GPR-OdomNet: Difference and Similarity-Driven Odometry Estimation Network for Ground Penetrating Radar-Based Localization\\

}

\author{
\IEEEauthorblockN{1\textsuperscript{st} Huaichao Wang}
\IEEEauthorblockA{\textit{the Department of Computer Science} \\
\textit{Civil Aviation University of China}\\
Tianjin, China \\
hc-wang@cauc.edu.cn}
\and
\IEEEauthorblockN{2\textsuperscript{nd} Xuanxin Fan}
\IEEEauthorblockA{\textit{the Department of Computer Science} \\
\textit{Civil Aviation University of China}\\
Tianjin, China \\
755824110@qq.com}
\and
\IEEEauthorblockN{3\textsuperscript{rd} Ji Liu}
\IEEEauthorblockA{\textit{Chengdu Textile College} \\
\textit{Chengdu Textile College}\\
Chengdu, China \\
liuji\_haha@126.com}
\and
\IEEEauthorblockN{4\textsuperscript{th} Dezhen Song}
\IEEEauthorblockA{\textit{the Department of Robotics} \\
\textit{Mohamed Bin Zayed University of Artificial Intelligence (MBZUAI)}\\
Abu Dhabi, UAE \\
dezhen.song@mbzuai.ac.ae}
\and
\IEEEauthorblockN{5\textsuperscript{th} Haifeng Li}
\IEEEauthorblockA{\textit{the Department of Computer Science} \\
\textit{Civil Aviation University of China}\\
Tianjin, China \\
hfli@cauc.edu.cn}
}

\maketitle

\begin{abstract}
When performing robot/vehicle localization using ground penetrating radar (GPR) to handle adverse weather and environmental conditions, existing techniques often struggle to accurately estimate distances when processing B-scan images with minor distinctions. This study introduces a new neural network-based odometry method that leverages the similarity and difference features of GPR B-scan images for precise estimation of the Euclidean distances traveled between the B-scan images. The new custom neural network extracts multi-scale features from B-scan images taken at consecutive moments and then determines the Euclidean distance traveled by analyzing the similarities and differences between these features. To evaluate our method, an ablation study and comparison experiments have been conducted using the publicly available CMU-GPR dataset. The experimental results show that our method consistently outperforms state-of-the-art counterparts in all tests. Specifically, our method achieves a root mean square error (RMSE), and achieves an overall weighted RMSE of 0.449 m across all data sets, which is a 10.2\% reduction in RMSE when compared to the best state-of-the-art method. 
\end{abstract}

\begin{IEEEkeywords}
Ground Penetrating Radar (GPR), Robot Localization, Deep Learning Odometry.
\end{IEEEkeywords}

\section{Introduction}

Precise localization of robots and vehicles in challenging weather and environmental scenarios is essential for autonomous driving. Popular localization methods rely on onboard sensors such as the global positioning system (GPS) receiver, cameras, lidars, and inertial measurement units (IMU)\cite{23,24,26}. However, these methods are significantly hampered in environments such as urban canyons, tunnels, or under adverse weather conditions, presenting significant safety challenges in autonomous driving. In contrast, the subsurface structures and features of an urban road remain stable and less affected by these adverse conditions. Ground penetrating radar (GPR), as a less-exploited sensor that is complementary to existing sensors, can effectively detect subsurface features, providing new opportunities to enhance the reliability and robustness of localization under challenging weather or environmental scenarios.

\begin{figure}[ht]
\centering
\includegraphics[scale=0.35]{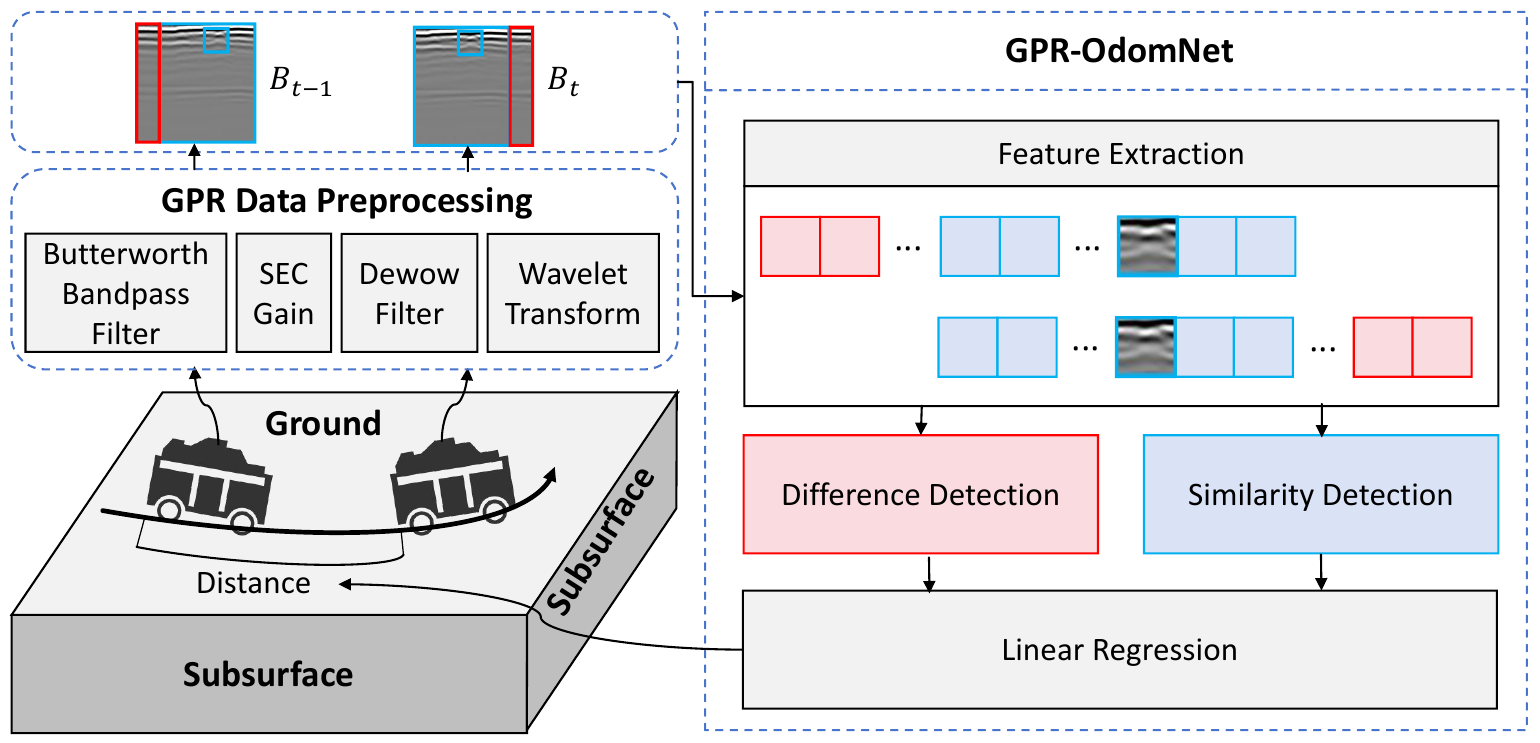}
\caption{A new GPR odometry network that estimates the distance by comparing the differences and similarities between two consecutive time step B-scan images.}
\label{fig:global}
\end{figure}

Here, we introduce a novel deep learning-based GPR odometry network (GPR-OdomNet) to capture the properties of subsurface features. As shown in Fig.~\ref{fig:global}, GPR-OdomNet uniquely exploits similarity and difference in consecutive GPR B-scan images, achieving more accurate distance estimation by capturing both high-level and subtle features. We conducted extensive experiments using the publicly available CMU-GPR dataset~\cite{17}. The experimental results show that our method consistently outperforms state-of-the-art counterparts in all tests. Specifically, our method reaches an overall absolute trajectory error (ATE) of 0.449 meters, which is measured in root mean square error (RMSE)  aggregated over all scenes. The result is a 10.2\% reduction in RMSE compared to the best state-of-the-art method. 

\section{Related Work}

The closely related work includes general GPR perception applications, GPR-based localization development, and recent deep learning-based approaches, in particular. 

Due to its unique sensing modality, GPR finds many application domains, including bridge inspection~\cite{1,2}, infrastructure assessment~\cite{3,4}, transportation infrastructure diagnostics~\cite{5,6}, and extraterrestrial exploration~\cite{7,8}. These applications primarily utilize GPR's subsurface target detection capabilities. When mounted on a robot, GPR enables 3D underground reconstruction through nondestructive scanning~\cite{9,10,11}, demonstrating GPR's potential for high-precision spatial mapping.

\begin{figure*}[ht]
\centering
\includegraphics[scale=0.6]{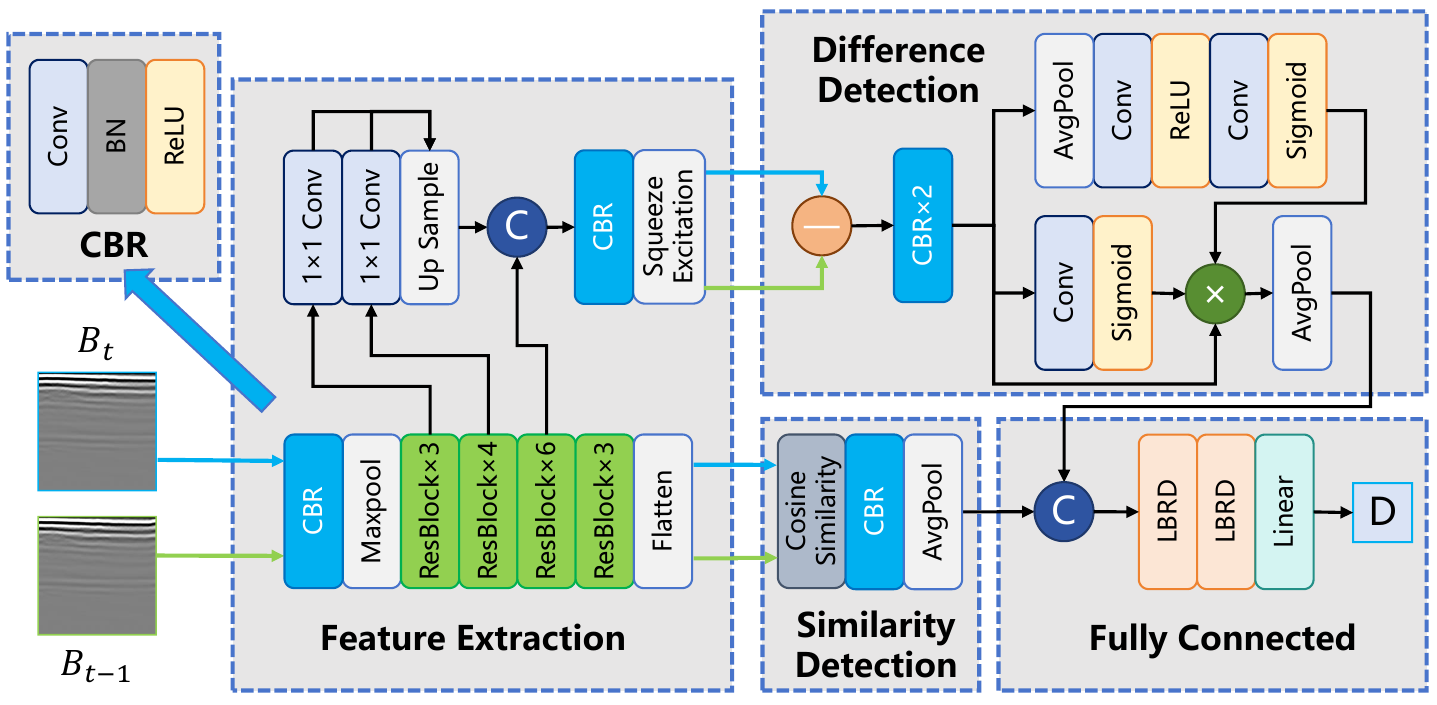}
\caption{System architecture of GPR-OdomNet. The network takes B-scan images from two consecutive time steps as input. Initially, multi-scale features are extracted from each image separately. Subsequently, the difference tensor is obtained by detecting difference between the two feature sets, and the similarity tensor is calculated through cosine similarity. Finally, the fully connected layers perform regression on the difference tensor and similarity tensor to yield the final distance. \textcircled{c} means tensor concatenation, \textcircled{×} means tensor multiplication, \textcircled{-} means tensor subtraction, and \fbox{$D$} means distance.}
\label{fig:program}
\end{figure*}
GPR-based localization research has gained a lot of research attention recently. The pioneering localization GPR (LGPR) system~\cite{12} proposes basic signal matching principles but suffers from raw signal noise, limited sensitivity, and GPS dependence. Skartados et al.~\cite{13} assume and utilize the existence of a widespread pipeline structure, which also limits the scope of practical application.  Li et al.~\cite{14} propose a Dominant Energy Curve (DEC) descriptor, enhancing accuracy via the metric feature mapping. Subsequent work~\cite{21} developed a multi-modal odometry system combining GPR with inertial and wheel sensors, introducing subsurface feature matrix (SFM) as a feature representation and employing factor graph-based optimization to form a sensor fusion-based approach that is more robust. Both of these methods are benefited from with hand-crafted features. However, such methods rely on expert-guided manual feature selection which introduces subjectivity and computational complexity.

Despite the progress in using current deep learning solutions with strong performance on GPR images with clear feature disparities, more information can be extracted from adjacent B-scans. Recognizing that precise displacement information is embedded in these subtle variations between the B-scans, our work designs a new neural network to capture the information, which can significantly advance localization accuracy and robustness.
\section{Problem Definition}

Before introducing the GPR localization problem, main variables are defined as follows,
\begin{itemize}
    \item $L$: the maximum width of the B-scan image, i.e., the number of sampling points in the time dimension.
    \item $B_t$: the B-scan data which is the set of A-scan data from time $t-L$ to time $t$.
    \item $O_{t-1,t}$: the traveled distance of the GPR between time $t-1$ and time $t$.
\end{itemize}
With notations defined, we formally define the problem as follows:

\begin{Pro}
Given $B_{t-1}$ and $B_{t}$, determine $O_{t-1,t}$.
\end{Pro}

\section{GPR-OdomNet}
We design a deep learning-based GPR odometry neural network (GPR-OdomNet) using the properties of subsurface features (see Fig.~\ref{fig:feature}). GPR-OdomNet takes inputs from the pre-processed B-scan images.
\subsection{GPR Data Preprocessing} 

We apply the same pre-processing steps as in~\cite{17,16} to raw GPR inputs to enhance signal quality and reduce noise, resulting in B-scan images for subsequent analysis. We skip the details and focus on the below primary steps for completeness purpose:
\begin{itemize}
\item A Butterworth bandpass filter is applied to eliminate high-frequency noise and low-frequency drift.
\item The spreading and exponential compensation (SEC) gain function is utilized to compensate for signal propagation loss.
\item Background noise is reduced using Dewow filter.
\item Wavelet transform is employed for denoising, thereby improving the signal-to-noise ratio.
\end{itemize}

\begin{figure}[htb!]
    \centering
    \begin{subfigure}[b]{0.48\textwidth}
        \centering
        \includegraphics[width=\textwidth]{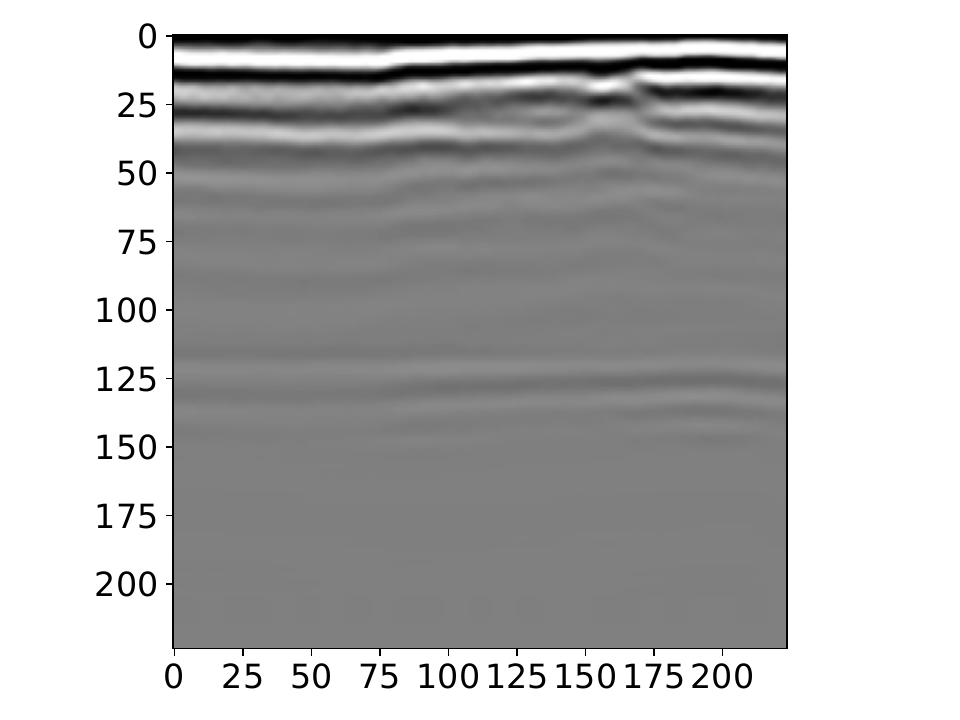}
        \caption{$B_{t-1}$}
        \label{fig:Bt-1}
    \end{subfigure}
    \hfill
    \begin{subfigure}[b]{0.48\textwidth}
        \centering
        \includegraphics[width=\textwidth]{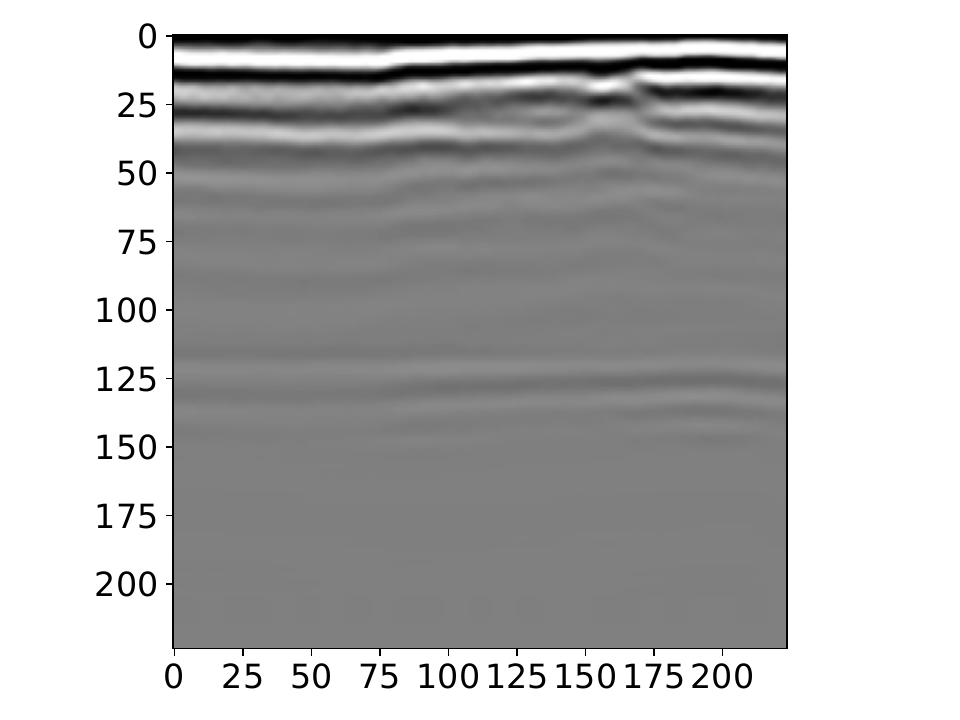}
        \caption{$B_t$}
        \label{fig:Bt}
    \end{subfigure}

    \vspace{0.5cm}

    \begin{subfigure}[b]{0.48\textwidth}
        \centering
        \includegraphics[width=\textwidth]{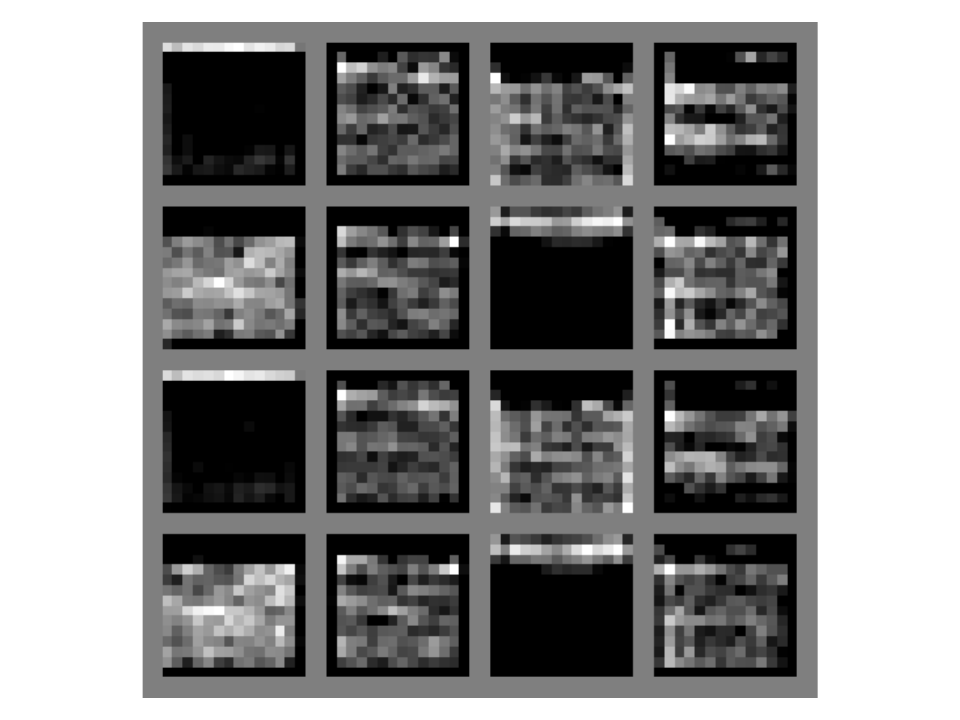}
        \caption{$F_{t-1,d}$ and $F_{t,d}$}
        \label{fig:Ftd}
    \end{subfigure}
    \hfill
    \begin{subfigure}[b]{0.48\textwidth}
        \centering
        \includegraphics[width=\textwidth]{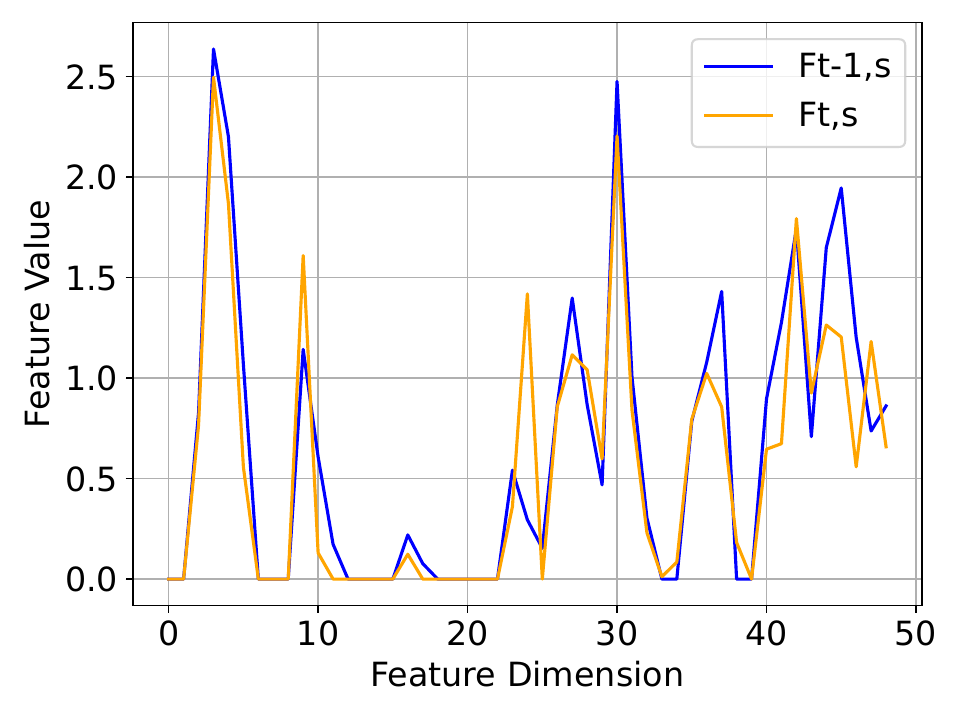}
        \caption{$F_{t-1,s}$ and $F_{t,s}$}
        \label{fig:Ft}
    \end{subfigure}

    \vspace{0.5cm}

    \begin{subfigure}[b]{0.48\textwidth}
        \centering
        \includegraphics[width=\textwidth]{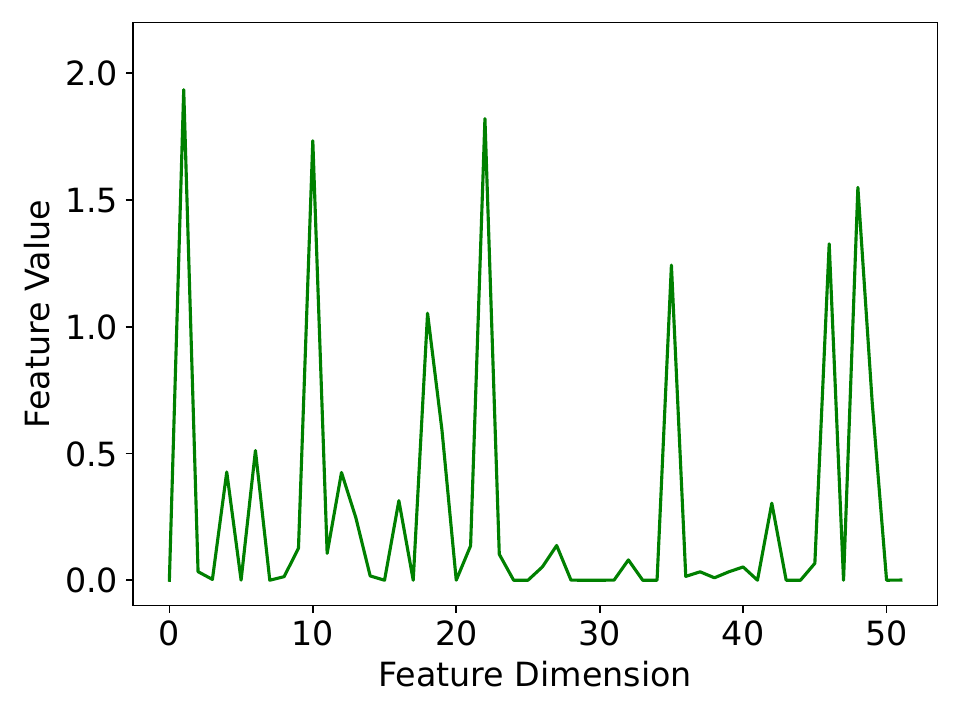}
        \caption{$D_{t-1,t}$}
        \label{fig:Difference}
    \end{subfigure}
    \hfill
    \begin{subfigure}[b]{0.48\textwidth}
        \centering
        \includegraphics[width=\textwidth]{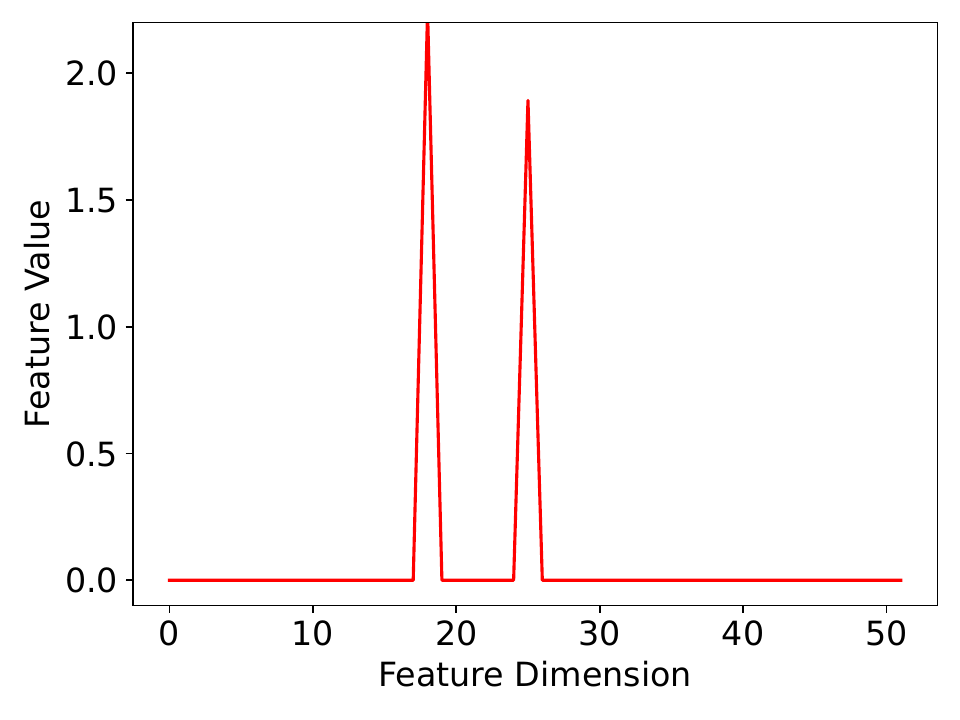}
        \caption{$S_{t-1,t}$}
        \label{fig:Similarity}
    \end{subfigure}
\caption{Examples of intermediate features of the GPR-OdomNet. Due to the large feature dimension of $F_{t-1,d}$, $F_{t,d}$, $F_{t-1,s}$, $F_{t,s}$, $D_{t-1,t}$ and $S_{t-1,t}$, the illustration here is a randomly sampled. The first eight subgraphs in (c) belong to $F_{t-1,d}$, while the last eight belong to $F_{t,d}$}
\label{fig:feature}
\end{figure}

Figs.~\ref{fig:feature}(a) and (b) illustrate two sample B-scan images after the pre-processing step through data interpolation and stitching, which serve as the input to the GPR-OdomNet.

\subsection{GPR-OdomNet}

The GPR-OdomNet estimates the distance traveled by the GPR using two consecutive B-scan images. This network is composed of four primary modules: feature extraction, difference detection, similarity detection, and fully connected regression, as depicted in Fig.~\ref{fig:program}.

\subsubsection{Feature Extraction}
Noise from the B-scan images is inevitable, which may obscure critical reflection signals. A key step is the suppression of noise and non-target reflections to enhance the visibility of effective reflection signals. we employ ResNet-50 with modified input channels for B-scan processing,
\begin{equation}\label{eq:input}
F_{t,1} = \mathcal{H}(B_t) + \mathcal{F}(B_t, W_l)
\end{equation}
where \( \mathcal{H}(\cdot) \) denotes identity mapping, \( \mathcal{F}(\cdot) \) represents residual transformations with weights \( W_l \), and \( F_{t,1} \) is a four-dimensional feature extracted by the first layer of ResNet-50(which includes 3 residual blocks, as shown in Fig.~\ref{fig:program}), including batch\_size, channels, width and height. 

After four layers of feature extraction using Resnet-50, we obtained multi-scale features \( F_{t,1} \), \( F_{t,2} \), \( F_{t,3} \) and \( F_{t,4}\). We compress  the low-dimensional features \( F_{t,1} \), \( F_{t,2} \) and \( F_{t,3} \) containing detailed information to obtain \( F_{t,d} \) for difference detection, and flatten the high-dimensional features \( F_{t,4}\) containing global information to obtain \( F_{t,s} \) for similarity detection. Figs.~\ref{fig:feature} (c) show some subgraphs of the extracted features $F_{t-1,d}$ and $F_{t,d}$. Figs.~\ref{fig:feature} (d) show the extracted features $F_{t-1,s}$ and $F_{t,s}$ that have been flattened. Due to the extracted feature dimension being 512, it is inconvenient to illustrate. Hence, the result shown in Figs.~\ref{fig:feature} (c) and (d) are randomly sampled for display. By the way, the result shown in Figs.~\ref{fig:feature} (e) and (f) are also randomly sampled for display. Such feature extraction not only mitigates the interference from noise, but also strengthens the feature expression of target structures within the images, thereby enhancing the accuracy of subsequent image analysis.

\subsubsection{Difference Detection} 
After extracting features \( F_{t,d} \), \( F_{t-1,d} \), \( F_{t,s} \) and \( F_{t-1,s} \) from the images \( B_t \) and \( B_{t-1} \), respectively, the next step is to use these four features for similarity and difference detection. Similarity detection identifies the common ground of the two features. However, adjacent consecutive B-scans are very similar in general, as shown in Figs.~\ref{fig:feature} (d). It is also important to identify their difference, which contains important information about the GPR motion. The absolute difference between the features of the two images is computed, resulting in a difference vector that represents the disparity in features,
\begin{equation}
\Delta F_{t-1,t} = |F_{t,d} - F_{t-1,d}| 
\end{equation}
Each element of $\Delta F_{t-1,t}$ reflects the magnitude of the discrepancy between the two images along the corresponding feature dimension. The absolute value is taken to disregard the direction of the difference, focusing solely on the magnitude of the disparity. 

To further assist in the extraction of local patterns and structural information from $\Delta F_{t-1,t}$, a convolutional operation is applied. The convolutional layer filters are adapted to capture local correlations within $\Delta F_{t-1,t}$, such as identifying whether the disparities along certain feature dimensions exhibit any spatial regularity. The introduction of an activation function, i.e.  ReLU, adds non-linearity to the network, enabling it to capture more complex patterns of feature differences,
\begin{equation}
\text{CBR}(x) = \text{ReLU}(\text{BN}(\text{Conv}(x)))
\end{equation}
\begin{equation}
\Delta F_{t-1,t}^\text{Conv} = \text{CBR}(\text{CBR}(\Delta F_{t-1,t}))
\end{equation}
where Conv stands for convolution and BN represents batch normalization.

To enhance the discriminative power of $\Delta F_{t-1,t}$, we have added spatial-channel attention mechanisms for the difference features after convolution. The spatial-channel attention module simultaneously captures both channel-wise interdependencies and spatial importance, enhancing the network's ability to focus on semantically significant features. The attention mechanism consists of two components:

\begin{itemize}
\item \textbf{Channel Attention:} Computes attention weights using global average pooling and multi-layer perceptron with sigmoid activation:
\begin{equation}
\begin{aligned}
\text{CA}(x) = \sigma\left(\text{Conv}\left(\text{ReLU}\left(\text{Conv}\left(\text{GAP}(x)\right)\right)\right)\right) \\
\end{aligned}
\end{equation}
where GAP denotes global average pooling, and $\sigma$ is the sigmoid function.

\item \textbf{Spatial Attention:} Computes spatial attention weights using spatial context aggregation with large receptive field:
\begin{equation}
\text{SA}(x) = \sigma\left(\text{Conv}(x)\right)
\end{equation}
\end{itemize}

The final attention-enhanced difference representation is obtained by applying both attention mechanisms sequentially. The enhanced difference features are then aggregated through global average pooling for subsequent processing:
\begin{equation}
D_{t-1,t} = \text{GAP}(\Delta F_{t-1,t}^\text{Conv} \otimes \text{CA}(\Delta F_{t-1,t}^\text{Conv}) \otimes \text{SA}(\Delta F_{t-1,t}^\text{Conv}))
\end{equation}
where $\otimes$ denotes tensor multiplication, and $D_{t-1,t}$ as shown in Fig.~\ref{fig:feature}(e) represents the feature vector capturing the differences between $F_{t-1,d}$ and $F_{t,d}$, describing the variations in GPR signals between adjacent time windows. This combined approach effectively captures both the magnitude and spatial-channel significance of feature differences between adjacent time windows.

\subsubsection{Similarity Detection} In parallel to the difference detection, the network evaluates the similarity features \( S_{t-1,t} \) between the two B-scan images to enhance the accuracy of distance estimation,
\begin{equation}
\text{CS}(x,y) = \frac{x \cdot y}{\|x\|_2 \cdot \|y\|_2}
\end{equation}
\begin{equation}
S_{t-1,t} = \text{GAP}(\text{CBR}(\text{CS}(F_{t-1,s},F_{t,s})))
\end{equation}
where ${S_{t-1,t}}$ (see Fig.~\ref{fig:feature}(f)) represents the feature vector describing the similarities between ${F_{t-1,s}}$ and ${F_{t,s}}$. And \(\|\cdot\|_2\) denotes the L2 norm (euclidean norm) of a vector, calculated as the square root of the sum of its squared components. Here, the cosine similarity (CS) between the extracted features measures the similarity of the two images in the feature space. The cosine similarity ranges from \(-1\) to \(1\), with a value closer to \(1\) indicating a higher similarity between the images in the feature space, suggesting a smaller distance and vise versa. Next, after convolution and pooling, overly similar features are tended towards zero to highlight salient features and reduce computational complexity. Comparing Fig.~\ref{fig:feature}(e) and Fig.~\ref{fig:feature}(f), as analyzed in the previous section, we can see that due to the excessive similarity between $F_t$ and $F_{t-1}$, the similarity detection result $S_{t-1,t}$ (see Fig.~\ref{fig:feature}(f)) tends to approach \(0\) in all dimensions, which reduces the effectiveness of subsequent regression. However, the difference detection results $D_{t-1,t}$ (see Fig.~\ref{fig:feature}(e)) have more significant feature dimensions, which provide a better basis for subsequent regression.

\subsubsection{Fully Connected Regression} 
Finally, the regression module estimates the distance by concatenating the difference features \( D_{t-1,t} \) and the similarity features \( S_{t-1,t} \),
\begin{equation}
    \text{LBRD}(x) = \text{Dropout}( \text{ReLU}( \text{BN}( \text{Linear}(x))))
\end{equation}
\begin{equation}
    {D}_{t-1,t} = \text{Linear}(\text{LBRD}(\text{LBRD}(\text{Concat}(D_{t-1,t},S_{t-1,t})))) 
\end{equation}
where \( {D}_{t-1,t} \) signifies the predicted value by the network for the \((t-1,t)\)-th distance, \text{Linear} represents fully connected layers, and \text{Concat} represents tensor concatenation.

The overall network is trained using the Root Mean Square Error (RMSE) function with supervised ground truth data obtained from a total station,
\begin{equation}
\label{eq:RMSE}
\mathcal{L} = \sqrt{\frac{1}{n}\sum_{t=1}^{T}(O_{t-1,t} - \hat{O}_{t-1,t})^2}
\end{equation}
where \( T \) denotes the total number of samples.

\section{Experiments}

\begin{figure}
\centering
\includegraphics[scale=0.45]{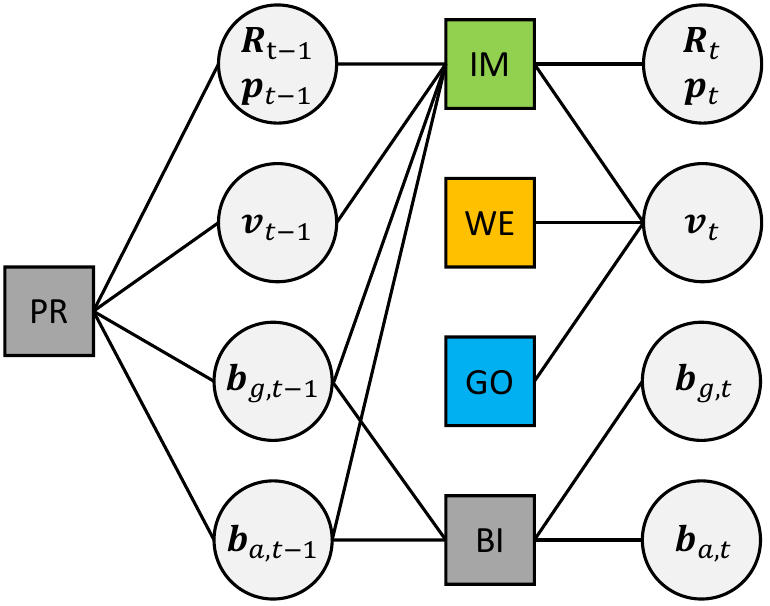}
\caption{Overview of the factor graph model. Square nodes represent factor nodes, and circular nodes represent variable nodes. The prior (PR) node is a prior factor node. The IMU (IM) node mainly constrains the pose and velocity between two consecutive moments. The Wheel Encoder (WE) and GPR-OdomNet (GO) nodes constrain the velocity of the next moment $t$. The Bias (BI) node constrains the biases between two consecutive moments.}
\label{fig:factor}
\end{figure}

We have implemented the proposed algorithm using Python 3.9 with packages such as torch 2.0.1, torchvision 0.15.2, and Pillow 9.5.0 on a workstation with an Intel Xeon Silver 4216 processor, 256 GB random access memory (RAM), and an NVIDIA GeForce RTX 3090 Graphics Processing Unit (GPU), running on an Operating System of Ubuntu 22.04.1. 
Firstly, we introduce the dataset used in the experiment. Secondly, we evaluate GPR-OdomNet based relative distance estimation when using adjacent GPR frames alone. Finally, we combine GPR-OdomNet with other sensory inputs in a full-fledged sensor-fusion-based odometry approach which allows us to compare overall performance to that of the state-of-the-art counterparts.

\subsection{Dataset}
We have evaluated our GPR-OdomNet using the widely accepted public CMU-GPR dataset\cite{17}, which is often used in GPR-assisted robot navigation. It covers three scenarios where GPS signals are unavailable, including four trajectories in the basement (\texttt{nsh\_b}), three trajectories in the factory workshop (\texttt{nsh\_h}), and seven trajectories in the parking lot (\texttt{gates\_g}). For these three environments, the total trajectory lengths of the \texttt{gates\_g}, \texttt{nsh\_b}, and \texttt{nsh\_h} datasets are 365, 264, and 90 meters, respectively. For each trajectory, the collected data include a single channel GPR, a camera, a Wheel Encoder, and a total station, with the total station measurements serving as the ground truth.

\subsection{Ablation Study}
To evaluate the effectiveness of each key module in GPR-OdomNet, we have performed an ablation study. Tab.~\ref{tab:ablation} shows the results of the performance comparison of different network variants, where the network performance is quantified using the RMSE error defined in~\eqref{eq:RMSE}. A lower RMSE value indicates a smaller deviation between the predicted distance and the ground truth distance from the total station. Specifically, \textit{Feature Concatenation} serves as the baseline method, employing a simple feature concatenation strategy; \textit{Difference Only} network focuses solely on image differences; \textit{Similarity Only } network considers only image similarity; and our full version of \textit{GPR-OdomNet}. 
\begin{table}[H]
\caption{RMSE (cm) of ablation studies}
\centering
\label{tab:ablation}
\begin{tabular}{cccc}
\toprule
\multirow{2}{*}{\textbf{Network Configuration}} & \multicolumn{3}{c}{\textbf{Dataset}}  \\
& gates\_g & nsh\_b & nsh\_h \\
\midrule
Feature Concatenation   & 11.574    & 7.936  & 5.636  \\
Similarity Only  & 5.214   & 3.007 & 2.802\\
Difference Only  & 3.749    & 1.912  & 2.195 \\
GPR-OdomNet   & \textbf{3.579}    & \textbf{1.808}    & \textbf{1.810}\\
\bottomrule
\end{tabular}
\end{table}
As shown in the table, GPR-OdomNet performs better than its degraded variants in all datasets, validating the need to jointly exploit difference detection and similarity detection. 
It should be noted that GPR-OdomNet reduces the average RMSE by 69.1\% compared to the worst-performing variant (\textit{Feature Concatenation}).
\subsection{Comparison of GPR-Only Relative Position Estimation Performance}
We then compare our \textit{GPR-OdomNet} with state-of-the-art GPR methods to estimate distance between adjacent frames. This is a non-cumulative relative position estimation from frame matching. 
During the experiment, we have classified the dataset based on the characteristics of trajectories (such as smooth circular or sharp turns): 4 trajectories from the \texttt{gates\_g} dataset are selected for training and 2 for testing; 3 trajectories from the \texttt{nsh\_b} dataset are used for training and 1 for testing; 2 trajectories from the \texttt{nsh\_h} dataset are used for training and 1 for testing. This partitioning method ensures the diversity and representativeness of the samples, providing a better comparison among all candidate methods.

Tab.~\ref{tab:compare} presents the odometry performance comparison between our method and three state-of-the-art GPR odometry methods. 
\begin{table}[H]
\caption{RMSE (cm) of different GPR-based relative distance estimation methods}
\centering
\setlength{\tabcolsep}{2pt}
\begin{tabular}{c|ccc|c}
\toprule
\multirow{2}{*}{Model} & \multicolumn{3}{c}{Dataset} &   \\
\cmidrule{2-5}
 & gates\_g & nsh\_b & nsh\_h & Overall \\
\midrule
Learned GPR Method\cite{16}   & 3.848    & 1.697  & 1.975  & 3.175\\
SFM\cite{21}  & 5.582  & 5.138  & 3.000  & 5.270\\
CNN-LSTM\cite{18}   & 5.975    & 3.528  & -  & -\\
GPR-OdomNet (Ours)   & 3.579    & 1.808    & 1.810  & \textbf{2.945}\\
\bottomrule
\end{tabular}
\label{tab:compare}
\end{table}
The \textit{CNN-LSTM} method has not been tested under the \texttt{nsh\_h} dataset in \cite{18}, and therefore, the corresponding entry is missing. It is clear that GPR-OdomNet and the \textit{Learned GPR Method} outperform \textit{SFM} and \textit{CNN-LSTM}. The GPR-OdomNet outperforms the \textit{Learned GPR Method} in two out of three datasets. Overall, our method reduces RMSE by 7.2\% when compared with the best state-of-the-art method \textit{Learned GPR Method}.

\subsection{Comparison of Odometry Performance in Sensor Fusion}

Since GPR odometry is often used in combination with other sensors such as IMU and Wheel Encoder, it is necessary to compare the localization performance in the sensor fusion setting.  This study designs a set of comparative experiments: a factor graph model in~\cite{21} that fuses the input of the IMU and the Wheel Encoder is used as the baseline method. Then GPR-OdomNet is incorporated using the actor graph model, as shown in Fig.~\ref{fig:factor}. Within the adopted factor graph optimization framework, the IMU observations are processed using a pre-integration method to constrain the robot's position, orientation, and velocity, while the measurements from the Wheel Encoder and the estimates from the GPR-OdomNet constrain the robot's velocity.

\begin{figure}
\centering
\includegraphics[scale=0.5]{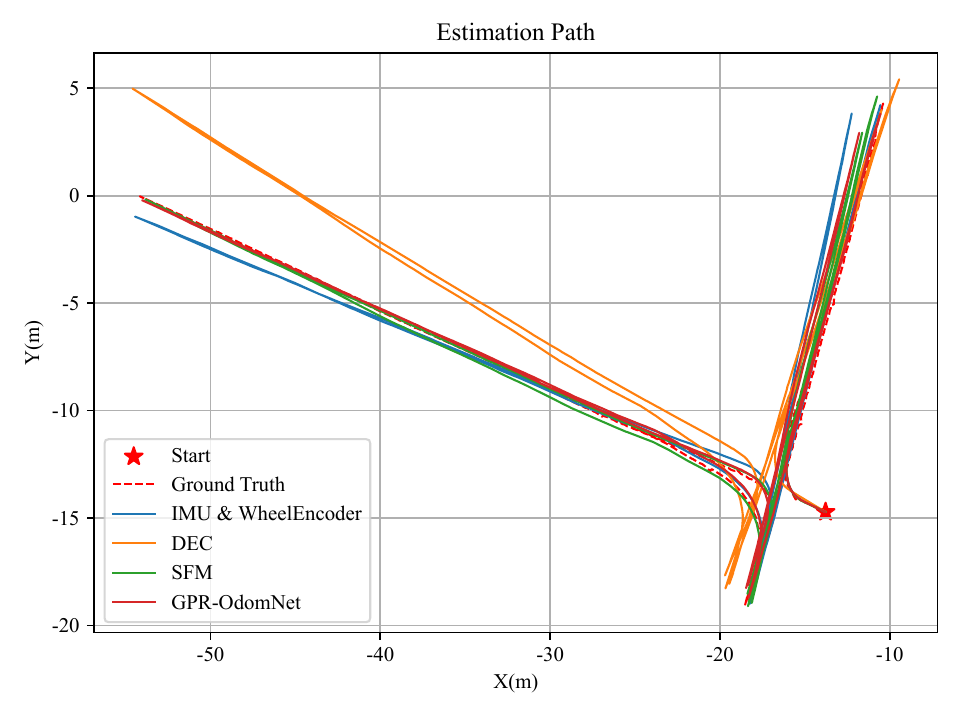}
\caption{Comparison of odometry performance in sensor fusion using the \texttt{gates\_g} dataset.}
\label{fig:path}
\end{figure}

The experimental results are shown in Fig.~\ref{fig:path}, where the \textit{Ground Truth} represents data collected by the aforementioned total station. The \textit{IMU \& WheelEncoder} does not include GPR input, which serves as the baseline method. The \textit{GPR-OdomNet} is our method here that employs GPR-OdomNet as the GPR component of the sensor fusion. The \textit{SFM} integrates GPR odometry extracted using a subsurface feature matrix (SFM)~\cite{21}. The \textit{DEC}~\cite{14} is an absolute localization method based on prior maps for DEC extraction and matching. The results show that the odometer accuracy of the \textit{GPR-OdomNet} model is superior to other models.

\begin{figure}
\centering
\includegraphics[scale=0.5]{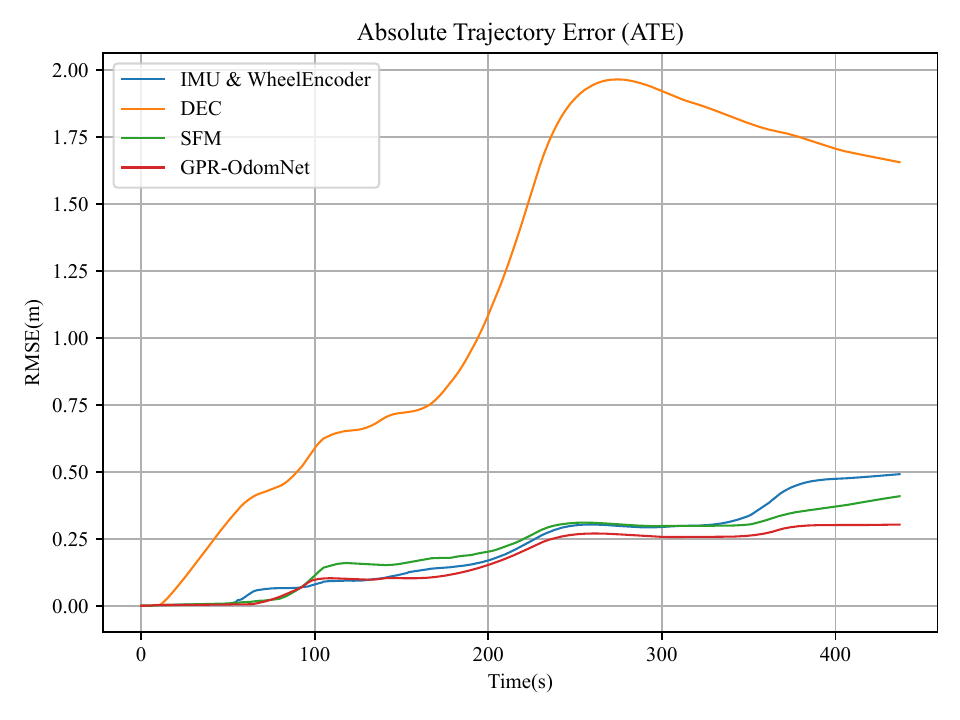}
\caption{Analysis of ATE RMSE over time corresponding to the trajectory in Fig.~\ref{fig:path}.}
\label{fig:ATE}
\end{figure}

To further compare odometry performance, Fig.~\ref{fig:ATE} presents RMSE values of Absolute Trajectory Error (ATE)~\cite{25} corresponding to the trajectories in Fig.~\ref{fig:path},
\begin{equation}
\mathcal{L}_{ATE} = \sqrt{\frac{1}{2T}\sum_{t=1}^{T}\left((x_t - \hat{x}_t)^2 + (y_t - \hat{y}_t)^2\right)}
\end{equation}
where $T$ represents the total duration of the robot's movement, $x_t$ and $y_t$ denote the horizontal and vertical coordinates obtained from the total station at time $t$, while $\hat{x}_t$ and $\hat{y}_t$ represent the estimated horizontal and vertical coordinates of the robot at time $t$. This metric provides a global assessment of the localization and odometry result, and hence is more comprehensive. It is clear that the \textit{GPR-OdomNet} method outperforms others.

Tab.~\ref{tab:ATE} details the RMSE for each dataset for each method. All methods employ the factor graph in Fig.~\ref{fig:factor} with the same input from IMU and Wheel Encoder. The only difference is whether or what type of GPR odometry is included. Among these, \textit{Learned GPR Method} integrates the learning sensor method~\cite{16}. \textit{Overall}, the last column in the Tab.~\ref{tab:ATE},  represents ATE weighted by trajectory length across all three datasets, 
\begin{equation}
\mathcal{L}_{Overall} = \frac{\sum_{i=1}^{n}W_i \cdot  \mathcal{L}_{ATE,i}}{\sum_{i=1}^{n}W_i}
\end{equation}
where $W_i$ denotes the total travel distance of the $i^{th}$ trajectory, and $\mathcal{L}_{ATE,i}$ represents the ATE RMSE of the $i^{th}$ trajectory calculated using~\eqref{eq:RMSE}. 

\begin{table}[htbp]
\centering
\caption{ATE RMSE (m) comparison results across datasets.}
\setlength{\tabcolsep}{2pt}
\begin{tabular}{c|ccc|c}
\toprule
\multirow{2}{*}{Model} & \multicolumn{3}{c}{Dataset} &   \\
\cmidrule{2-5}
 & gates\_g & nsh\_b & nsh\_h & Overall \\
\midrule
Learned GPR Method  & 1.228  & 0.332  & 0.251  & 0.590  \\
SFM  & 0.468  & 0.734  & 0.439  & 0.568  \\
DEC    &0.47   &0.52   &0.57   &0.50   \\
IMU \& WheelEncoder & 0.588    & 1.074  & 1.015  & 0.743 \\
GPR-OdomNet  & 0.353    & 0.751  & 0.380  & \textbf{0.449}  \\
\bottomrule
\end{tabular}
\label{tab:ATE}
\end{table}
The experimental results of the method \textit{DEC} are reported at two decimal places. Tab.~\ref{tab:ATE} show that the proposed \textit{GPR-OdomNet} achieves the best overall performance. For specific datasets, it performs best in the largest data set \texttt{gates\_g} and the second best in dataset  \texttt{nsh\_h}. However, the performance on dataset \texttt{nsh\_b} is average, which maybe due to the higher proportion of regular steel bars and pipes in this dataset. Further study is needed to analyze the issue. Nevertheless, the overall performance show that our method reduce ATE RMSE by at least by 10.2\% when compared with the best state-of-the-art method \textit{DEC}. It is also worth noting that all methods with GPR inputs outperforms the baseline \textit{No GPR} which means using GPR is necessary and helpful in odometry.

\section{Conclusion and Future Work}

A novel GPR odometry network based on 2D radar scan image that extracts difference and similarity feature was presented in the paper. The network estimated GPR traveling distance by exploiting feature differences and similarity information. Through systematic ablation experiments and comparative experiments conducted on the publicly available and widely-used CMU-GPR datasets, the results show that the proposed network improved over the state-of-the-art method by reducing odometry RMSE by at least 10.2\%. 

In the future, we will investigate deep learning-based multi-sensor fusion methods to construct an end-to-end integrated positioning system. This will further enhance the practicality and robustness of the model.

\bibliography{references}

\end{document}